\documentclass[conference]{IEEEtran}
\IEEEoverridecommandlockouts
\usepackage{times}
\usepackage{amsmath,amssymb,amsfonts}
\usepackage{algorithm}
\usepackage{algpseudocode}
\usepackage{xcolor}

\usepackage[numbers]{natbib}
\usepackage{multicol}
\usepackage[bookmarks=true]{hyperref}
\usepackage{graphicx}
\usepackage{booktabs}

\title{
    \LARGE \bf Vote-Tree-Planner: Optimizing Execution Order in LLM-based Task Planning Pipeline via Voting
}

\author{
    \IEEEauthorblockN{Chaoyuan Zhang$^*$, Zhaowei Li$^*$, Wentao Yuan$^{\dagger}$} 
    \IEEEauthorblockA{
        University of Washington \\
        \{cz86, lzw365, wentaoy\}@uw.edu
    }
    \thanks{$^*$ Equal contribution in random order.}
    \thanks{$^\dagger$ Corresponding author.}
}

\begin{document}


\maketitle

\begin{abstract}
Integrating large language models (LLMs) into closed-loop robotic task planning has become increasingly popular within embodied artificial intelligence. Previous efforts mainly focused on leveraging the strong reasoning abilities of LLMs to enhance task planning performance while often overlooking task planning efficiency and executability due to repetitive queries to LLMs. This paper addresses the synergy between LLMs and task planning systems, aiming to minimize redundancy while enhancing planning effectiveness. Specifically, building upon Prog-Prompt and the high-level concept of Tree-Planner, we propose Vote-Tree-Planner. This sampling strategy utilizes votes to guide plan traversal during the decision-making process. Our approach is motivated by a straightforward observation: assigning weights to agents during decision-making enables the evaluation of critical paths before execution. With this simple vote-tree construction, our method further improves the success rate and reduces the number of queries to LLMs. The experimental results highlight that our Vote-Tree-Planner demonstrates greater stability and shows a higher average success rate and goal condition recall on the unseen dataset compared with previous baseline methods. These findings underscore the potential of the Vote-Tree-Planner to enhance planning accuracy, reliability, and efficiency in LLM-based planning systems. 

\end{abstract}

\section{Introduction}
According to \citet{Kaelbling2011}, task planning is a crucial area in robotics, focusing on designing systems that construct sequences of mid-level actions to enable robots to perform complex high-level tasks. Effective task planning involves considering various factors, including robot capabilities, environmental variables, and potential constraints or uncertainties. A significant trend in this field, as highlighted by \citet{huang2022language} and \citet{song2023llmplanner}, is the use of Large Language Models (LLMs) to directly generate robotic actions, marking a departure from traditional methods that depend on predefined domains, such as those proposed by \citet{eysenbach2019search} and \citet{xu2019regression}.

The application of LLMs in planning has garnered considerable attention within the robotics community, motivated by both the demonstrated capabilities of AI systems to reason about complex scenarios and the demand from downstream applications, such as goal-driven robotics \cite{huang2022language} and intelligent planning assistants \cite{lyu2021goaloriented}. The most common approach involves employing LLMs as planners to generate action sequences leading to predefined goal states \cite{stein2024autoplanbench,valmeekam2023planbench}. However, despite its broad applicability, this LLM-based approach faces limitations, especially in text-simulated environments where it often underperforms and lacks the interpretability offered by symbolic planning methods that generate plans from formal environment representations.

To address the issue of interpretability, \citet{singh2022progprompt} introduced Prog-Prompt, a new prompting mechanism that leverages the logistic-rich nature of programming and LLMs' extensive knowledge of online code bases to enhance the interpretability of environmental textual representations. Although Prog-Prompt has significantly improved interpretability, it still faces challenges, such as repetitive commands and misinterpretations of textual representations, and offers limited options for correcting unsuccessful executions. Recent efforts by \citet{hu2023treeplanner} have attempted to resolve these issues by employing a tree-like structure to aggregate generated plans, enhancing the execution process through an LLM-based model. Nonetheless, this approach reintroduces interpretability challenges as it does not utilize a programming language-based prompting mechanism. Additionally, our experiments have shown that simply interacting with the environment to determine the success of an execution command can yield results comparable to those obtained by providing the LLM with continuous observations at each command.

In this paper, we introduce the Vote-Tree-Planner, a novel planning mechanism that combines the strengths of Prog-Prompt and the high-level concept of Tree-Planner to enhance the executability and reliability of plans generated by LLMs. Our approach employs a planning tree that adapts to unexpected situations and ensures consistent task execution. Experimental results demonstrate significant improvements in plan generation accuracy, reliability, and efficiency, underscoring the potential of Vote-Tree-Planner to advance the field of mid-level robotic task planning.

\begin{figure*}[ht]
    \centering
    \includegraphics[width=\textwidth, page=1]{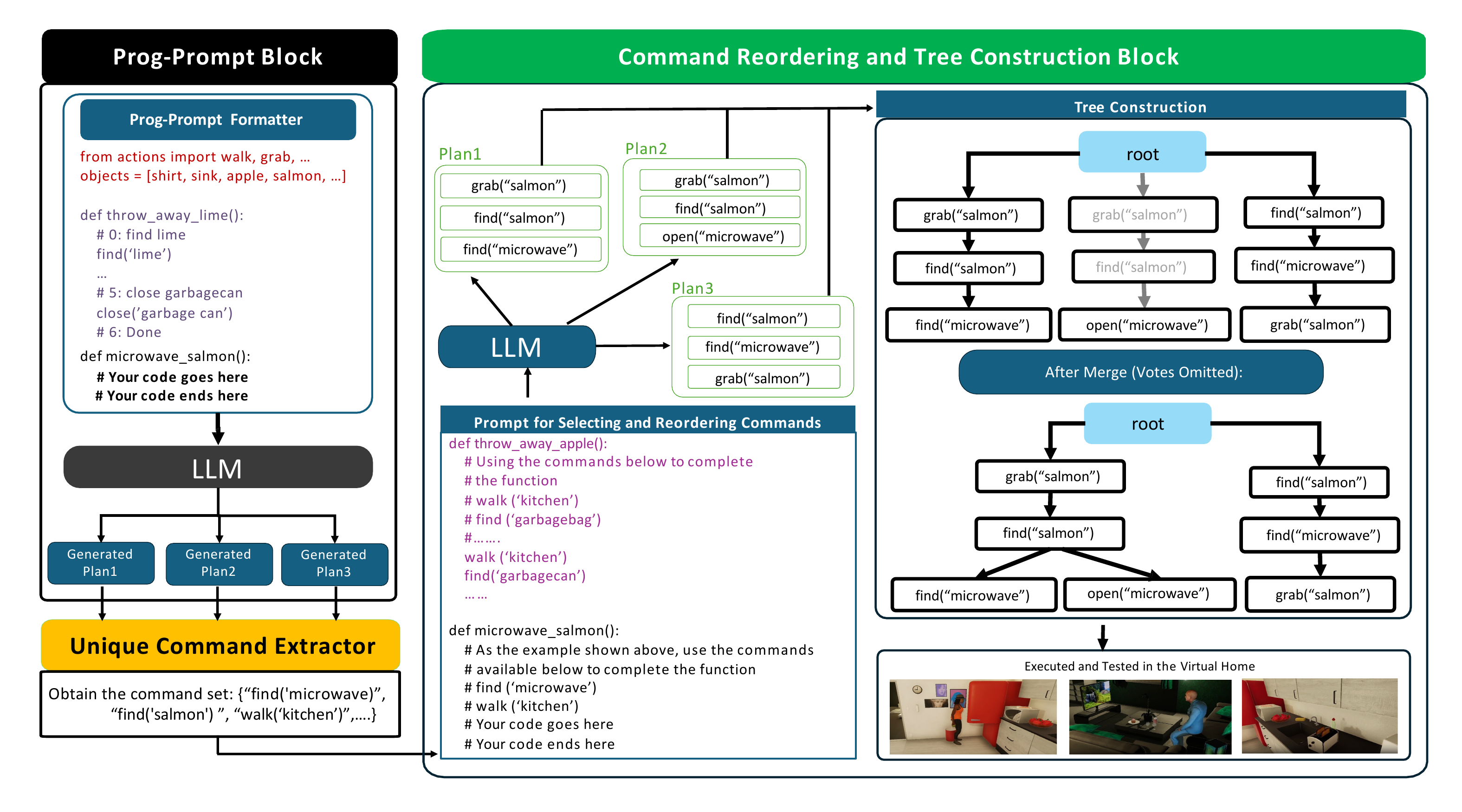}
    \caption{\textbf{Illustration of our proposed Vote-Tree-Planner pipeline}. $N$ plans are generated by the framework of Prog-Prompt using the LLM, then the unique commands are extracted and reordered into $M$ plans by the LLM again. These plans are then merged into an execution tree as inspired by Tree-Planner \cite{hu2023treeplanner}. The execution tree is then tested in the virtual home simulator \cite{puig2018virtualhome, puig2020watchandhelp}. We integrate the main characteristics from both pipelines to form our proposed method. Left (the Prog-Prompt Block): prompts the LLM and generates $N$ plans, then unique commands are extracted to a set. Middle (left column of the Command Reordering and Tree Construction Block): prompts the LLM again using the unique command set and generates $M$ plans. Right (right column of the Command Reordering and Tree Construction Block): constructs the generated plans into a tree with votes, and then executes the plan according to the votes.}
    \label{fig:pipeline}
\end{figure*}

\section{Related Work}
Task Planning is a crucial process in robotics, where robots generate a sequence of actions to complete tasks within specific environments \cite{Kaelbling2011}. Traditionally, Task Planning relies on heuristics and searches within predefined domains \cite{garrett2020pddlstream, jiang2019task}, with some studies exploring representation learning, hierarchical learning, and other methodologies \cite{eysenbach2019search, xu2018neural, zhang2023cat}. Recently, the development of Large Language Models \cite{brown2020language, chen2021evaluating} (LLMs) has initiated a shift towards leveraging these models to directly generate plans and facilitate correct executions due to their strong generation and reasoning capabilities \cite{chu2023timebench, gramopadhye2023generating, valmeekam2023planbench}.

The field of Task Planning has undergone significant evolution with the integration of LLMs, particularly in the domain of closed-loop planning. A seminal contribution in this area is Prog-Prompt \cite{singh2022progprompt}, which utilizes structured programming prompts to guide the planning process and leverages LLMs' common-sense knowledge for strategy adjustments during execution. By incorporating assertions in programming languages, Prog-Prompt enables robots to proactively gather environmental data, enhancing the precision and context-awareness of task planning. Unlike Prog-Prompt, the Tree-Planner does not employ a Python-like prompt mechanism but instead utilizes a tree-like structure to aggregate multiple generated plans \cite{hu2023treeplanner}. In its execution process, Tree-Planner uses LLMs as heuristics to select the next command based on current observations. Although their methods achieve outstanding performance, they lack specific designs to address redundant execution queries, which hampers the system's efficiency and stability. Compared with the aforementioned methods, our Vote-Tree-Planner incorporates the unique command extractor to minimize redundant commands and implement a voting mechanism to enhance the stability of generated plans and reduce the number of queries sent to the LLM.

\section{Problem Definition}
Given a high-level instruction such as ``Microwave Salmon'', our objective is to enable a robot to decompose this instruction into several intermediate commands, each representing a combination of predefined actions and objects. This decomposition process is designed to translate abstract commands into actionable sequences that the robot can execute effectively. We formalize the problem as the tuple \( \langle I, S, A, O, g, i \rangle \), where \( I \) represents a high-level instruction that describes a task the agent must complete, \( S \) is a set of states each describing the environment's state, \( A \) is a set of actions available to the agent to interact with the environment and manipulate objects, \( O \) is a set of objects within the environment, the relationships among which can be altered by actions \( a \in A \), and \( g \) and \( i \) are specific states in \( S \) representing the goal state and the initial state, respectively. This formulation captures the dynamic interactions between the agent and its environment, crucial for understanding and executing the given tasks effectively.

\section{Proposed Method: Vote-Tree-Planner}
To enable a robot to execute an abstract high-level instruction like ``Microwave Salmon'', this instruction must be converted into a plan composed of several executable, mid-level commands. We formalize our Vote-Tree-Planner as a planning sampling process:
\[
\text{Vote\_Tree\_Planner}(I, A, O) = \{a_1(o_1), \ldots, a_N(o_N)\},
\]
where each \(a_i \in A\) and \(o_i \in O\), $\forall$ \(i \in \{1, \ldots, N\}\).

We formulate the entire plan sampling process as the following stages, as shown in Figure \ref{fig:pipeline}. 1) \textbf{Prog-Prompt Formatting}: This stage entails converting the high-level instruction \(I\), along with available actions \(A\) and objects \(O\), into a structured Prog-Prompt \cite{singh2022progprompt} format. 2) \textbf{Plan Generation and Unique Command Extraction}: in this stage, a Large Language Model (LLM) is utilized to generate multiple potential plans using the formatted prompt. Then we extract unique commands in which each contains one action and one object. 3) \textbf{Reordering Prompt Formatting and Sequence Generation}: this stage entails the LLM reordering the extracted unique commands into a sequential plan that satisfies the initial instruction. 4) \textbf{Planning Tree Construction}: during the final stage, the new plans are structured into a tree-like format with votes in each node to enhance decision-making and execution efficiency. In the following sections, we will address each aforementioned component in detail.

\subsection{Prog-Prompt Formatting}
To enable a large language model (LLM) to transform a high-level instruction into a detailed action plan, we translate the instruction into the Prog-Prompt format as proposed by Singh et al. \cite{singh2022progprompt}. This translation process is formalized as
\[\rho_{\text{prog}} = \text{FORMATTER\_PROG}(I),\]
where \( \rho_{\text{prog}} \) represents the Prog-Prompt format, which contains necessary information of the instruction \(I\), available actions \(A\), and accessible objects \(O\). Figure \ref{fig:pipeline} (Left) illustrates a typical example of a Prog-Prompt formatted prompt.

\begin{figure*}
    \centering
    \includegraphics[width=\textwidth]{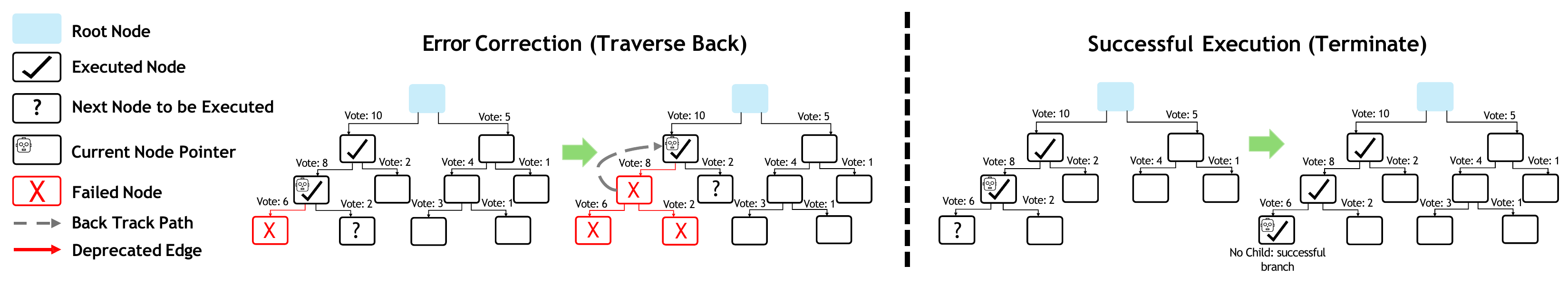}
    \caption{\textbf{Illustration of voting mechanism}. \textit{Error Correction} (left): When a node is successfully executed. The next step is finding the child of that node with the highest vote and trying to execute the command of that node. If the execution fails, we execute the second-highest-voted child. If all the children fail, the mechanism should traverse another branch according to the number of votes and continue. \textit{Successful Execution} (right): The execution process terminates when a node without any child nodes is executed successfully.}
    \label{fig:execution}
\end{figure*}

\subsection{Plan Generation and Unique Command Extraction}
The $\rho_{\text{prog}}$ from the last subsection is used to generate multiple executable plans. However, unlike traditional approaches \cite{ahn2022i, hu2023treeplanner, huang2022language, li2022pretrained, singh2022progprompt}, our methodology does not directly execute these generated plans at the first time. Instead, we extract unique commands and discard extraneous information. This strategy effectively narrows down the range of potential commands from all conceivable action-object combinations to only those that are specific and relevant, thereby reducing the risk of errors due to irrelevant or impractical combinations.

During the unique command extracting stage, our objective is to identify a set of commands, $\mathbb{S}=$ \( \{c_1, c_2, \ldots, c_N\} \), each command \( c_i \) containing one action \( a \in A \) as well as one or two associated objects \( (o_1) | (o_1, o_2) \in O \) depending on $a$. Each command in the set corresponds to commands ever appearing in the generated plans, focusing the selection process on combinations of elements from \( O \) and \( A \) that are most pertinent for completing the given task.

As illustrated in Figure \ref{fig:pipeline} (Left), a Prog-Prompt formatted prompt is fed into an LLM, which independently generates multiple plans. These plans are processed by the unique command extractor, which isolates distinct commands from each plan. This methodology not only refines the choices of action-object combinations but also ensures comprehensive coverage of all necessary combinations for task completion. The process can be formalized as follows: 
\[\text{COMMAND\_EXTRACTOR}(\text{LLM}(\rho_{\text{prog}}))\]\[ = \mathbb{S} = \{c_1, c_2, \ldots, c_N\}, \]
This approach refines the action-object combinations and extracts the essential commands necessary for task completion, distinguishing our method from those that directly utilize the outputs of LLMs as executable plans.
\subsection{Reordering Prompt Formatting and Sequence Generation}
After obtaining the set $\mathbb{S}$, our goal is to reorganize these commands into executable plans. As depicted in Figure \ref{fig:pipeline} (Middle), within the Command Reordering and Tree Construction Block, we integrate the commands from $\mathbb{S}$ into a reordering prompt alongside the original instruction \( I \). This section includes several examples that demonstrate how to achieve the given task by reorganizing the provided commands. This process is formalized as follows: 
\[ \text{COMMAND\_REORDER\_PROMPT}(\mathbb{S}, I) = \rho_{\text{reorder}}, \]
where \( \rho_{\text{reorder}} \) signifies the command reordering prompt. This prompt is subsequently input into a large language model (LLM) again. Through this process, denoted as: \[ \text{LLM}(\rho_{\text{reorder}}) = \mathbb{P} = \{p_1, p_2, \ldots, p_N\}, \] where \( \mathbb{P} \) represents the set of plans independently generated.

\begin{algorithm}
\caption{Construct Vote-Tree from Plans}
\begin{algorithmic}[1]
\State \textbf{Input:} Collection of reordered plans $\mathbb{P}$
\State \textbf{Output:} Vote-Tree $T$
\Procedure{BuildVoteTree}{$\mathbb{P}$}
    \State Initialize empty tree $T$
    \State Create root node $root$ of $T$ with no commands
    \For{each plan $p$ in $\mathbb{P}$}
        \State $currentNode \gets root$
        \For{each command $c$ in $p$}
            \If{$c$ is not in $currentNode.children$}
                \State Add a new child with $c$ to $currentNode$
            \EndIf
            \State $currentNode \gets currentNode.children[c]$
            \State $currentNode.vote++$
        \EndFor
    \EndFor
    \State \textbf{return} $T$
\EndProcedure
\end{algorithmic}
\label{algorithm:constructTree}
\end{algorithm}

\subsection{Vote-Tree Construction}
To select the optimal plan from the set \( \mathbb{P} \), directly executing each plan would be a straightforward method but proves to be time-consuming. Two primary issues arise: first, different plans may contain repetitive parts; second, some plans, due to the randomness inherent in the large language model (LLM) generation process, are not executable. To address these challenges, we have designed a tree-like structure that aggregates potential plans. As depicted in Figure \ref{fig:pipeline} (Right), plans sharing a common prefix are aggregated into a single branch. During this aggregation process, the number of plans combined at each node is counted and recorded as the variable \( \text{Vote} \), illustrated in Figure \ref{fig:execution}. Since plans may begin with different commands, the root node of the tree-like structure is not associated with any specific action. Algorithm \ref{algorithm:constructTree} details the process of constructing the Vote-Tree.

\begin{algorithm}
\caption{Execution Using Vote-Tree-Planner}
\begin{algorithmic}[1]
\State \textbf{Input:} Root node of the Vote-Tree
\State \textbf{Output:} Execution status

\Procedure{ExecutePlan}{$root$}
    \State $node \gets root$
    \While{not empty($node.children$)}
        \State $child \gets \Call{GetChildMaxVote}{node.children}$
        \State Execute command $c$ in $child$
        \If{Execution failed}
            \State remove $child$ from $node.children$
            \If{not empty($node.children$)}
                \State \textbf{continue}
            \Else
                \State $executed \gets node$
                \State $node \gets \Call{GetParent}{node}$
                \State remove $executed$ from $node.children$
            \EndIf
        \Else
            \State $node \gets child$
        \EndIf
    \EndWhile
\EndProcedure
\end{algorithmic}
\label{algorithm:executeTreePlanner}
\end{algorithm}

\subsection{Execution}
During the execution process, our objective is to consistently select the most optimal choice from the available actions. The variable \( \text{Vote} \), stored in each node of our tree structure, serves as a metric to indicate the most favorable command based on the large language model's (LLM's) preferences. If an execution attempt fails, the Vote-Tree-Planner will shift focus to the subsequent child node with the second highest \( \text{Vote} \). In scenarios where the executions of all children of a node fail, the planner will either terminate the process or, if feasible, backtrack to the last successful node that still possesses unexecuted child nodes. This dynamic decision-making process is depicted in Figure \ref{fig:execution} and Algorithm \ref{algorithm:executeTreePlanner}.

\section{Experiment}

\subsection{Experimental Setup}
\subsubsection{Environment}
Our experiments are conducted in the Virtual Home \cite{puig2018virtualhome, puig2020watchandhelp} environment, a robotic task simulation tool for common household tasks. Each Virtual Home scene includes hundreds of objects with individual properties and inter-object relationships. There are 28 different actions in the Virtual Home environment. Task-relevant goal conditions specify certain object states or predicates between objects, such as ``LIGHT is OFF'' for the action ``Turn Off Light'' and ``SALMON is in MICROWAVE'' for the action ``Put Salmon in Microwave''.

\subsubsection{Dataset}
The dataset we used is consistent with Prog-Prompt \cite{singh2022progprompt} and Tree-Planner \cite{hu2023treeplanner}. It contains 4 Virtual Home scenes and 35 unique Virtual Home tasks, each with a task name, goal conditions, and a goal plan. We generated the goal condition using the goal plan. 

\begin{table}
    \renewcommand{\arraystretch}{1.1}
    \caption{Experiment results without error correction.}
    \label{tab:without_correction_table}
    \resizebox{\linewidth}{!}
    {
    \begin{tabular}{c|c|c|c}
        \toprule[1.0pt]
        Methods & SR $\uparrow$ & GCR $\uparrow$ & Exec $\uparrow$ \\
        \hline
        Zero-Shot Planner \cite{huang2022language} & 0.01\tiny$\pm$0.01 & 0.02\tiny$\pm$0.01 & 0.16\tiny$\pm $0.03 \\
        \hline
        Prog-Prompt \cite{singh2022progprompt} & 0.34\tiny$\pm$0.06 & 0.67\tiny$\pm$0.07& \textbf{0.92}\tiny$\pm$0.02 \\
        \hline
        Tree-Planner \cite{hu2023treeplanner} & 0.28\tiny$\pm$0.02& 0.40\tiny$\pm$0.01& 0.55\tiny$\pm$0.01\\
        \hline
        Vote-Tree-Planner (Ours) & \textbf{0.43}\tiny$\pm$0.04 & \textbf{0.70}\tiny$\pm$0.04 & 0.89\tiny$\pm$0.02 \\
        \bottomrule[1.0pt]
    
    \end{tabular}}
\end{table}

\begin{table}
    \renewcommand{\arraystretch}{1.1}
    \caption{Experiment results with error correction.}
    \label{tab:with_correction_table}
    \resizebox{\linewidth}{!}
    {
    \begin{tabular}{c|c|c|c}
        \toprule[1.0pt]
        Methods & SR $\uparrow$ & GCR $\uparrow$ & Exec $\uparrow$ \\
        \hline
        Iterative-Planner \cite{hu2023treeplanner}\tiny{(Global)} & 0.37\tiny$\pm$0.02 & 0.52\tiny$\pm$0.01& 0.82\tiny$\pm$0.02 \\
        \hline
        Prog-Prompt \cite{singh2022progprompt} & 0.38\tiny$\pm$0.07 & 0.66\tiny$\pm$0.08& \textbf{0.93}\tiny$\pm$0.04 \\
        \hline
        Tree-Planner \cite{hu2023treeplanner} & 0.41\tiny$\pm$0.03& 0.60\tiny$\pm$0.03& 0.88\tiny$\pm$0.03\\
        \hline
        Vote-Tree-Planner (Ours) & \textbf{0.48}\tiny$\pm$0.07 & \textbf{0.81}\tiny$\pm$0.06 & 0.90\tiny$\pm$0.04 \\
        \bottomrule[1.0pt]
    
    \end{tabular}}
\end{table}

\begin{table}
    \renewcommand{\arraystretch}{1.1} 
    \caption{Experiment results (without error correction) of different node selection methods in Vote-Tree-Planner.}
    \label{tab:abalation_table}
    \resizebox{\linewidth}{!}{
    \begin{tabular}{c|c|c|c}
        \toprule[1.0pt]
        Node Selection Method & SR $\uparrow$ & GCR $\uparrow$ & Exec $\uparrow$ \\
        \hline
        Randomly Selected Node& 0.33\tiny$\pm$0.08 & 0.62\tiny$\pm$0.05 & 0.83\tiny$\pm$0.04 \\
        \hline
        Maximum Voted Node& \textbf{0.43}\tiny$\pm$0.04 & \textbf{0.70}\tiny$\pm$0.04 & \textbf{0.89}\tiny$\pm$0.02 \\
        \bottomrule[1.0pt]
    \end{tabular}}
\end{table}

\begin{figure*}
    \centering
    \includegraphics[width=\textwidth]{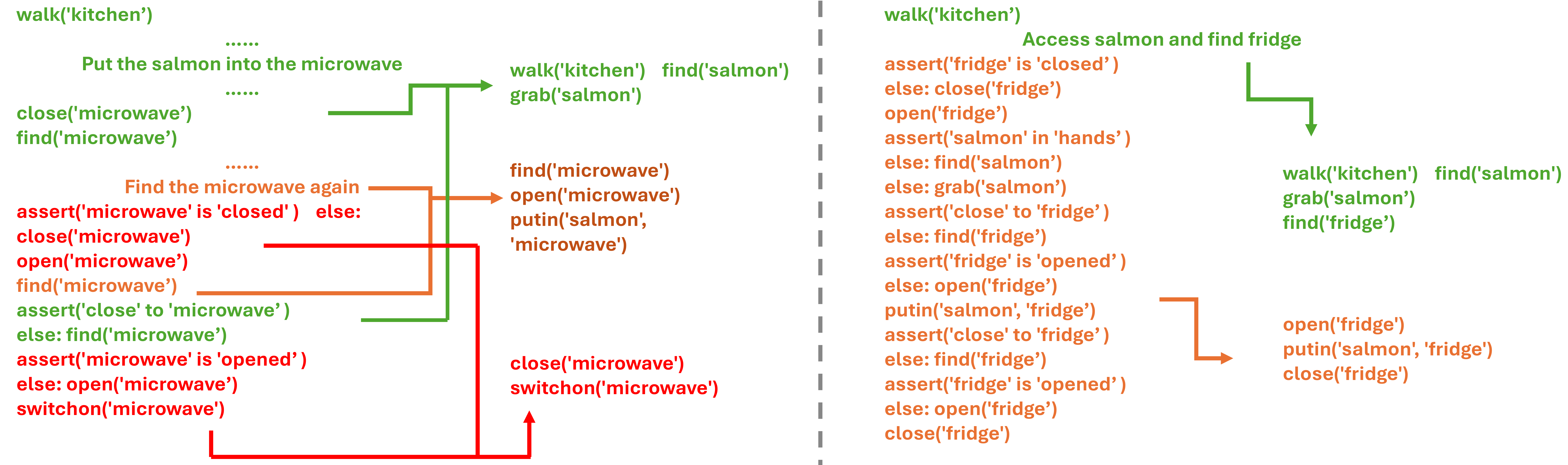}
    \caption{\textbf{Qualitative analysis}. The two planning examples, ``Microwave Salmon'' (left) and ``Put Salmon In The Fridge'' (right), show the comparison between two plans generated by Prog-Prompt \cite{singh2022progprompt} (on the left of each side) and our method (on the right of each side). Commands in \textcolor{red}{red} represent what Prog-Prompt did wrong and redundantly, while our method did correctly and concisely. Commands in \textcolor{orange}{orange} represent what Prog-Prompt did redundantly while our method did concisely. Commands in \textcolor{green}{green} represent necessary and correct commands for both Prog-Prompt and our method.}
    \label{fig:comparison}
\end{figure*}

\subsubsection{Evaluation Metrics}
Following the evaluation metrics in previous works \cite{hu2023treeplanner, singh2022progprompt}, we used success rate (SR), goal conditions recall (GCR), and executability (EXEC) as the main metrics to evaluate our pipeline performance. Specifically, the goal conditions are the set difference between final states and initial states during the execution. GCR is calculated by one minus the quotient of dividing the set difference between the ground truth final state goal conditions ($g$) and the achieved final state goal conditions ($g'$) by the total number of ground truth final state goal conditions. In other words, GCR represents the percentage of goal conditions achieved. SR is the fraction of tasks that achieve all goal conditions. Each task achieves an SR of 1 only if its GCR equals 1.
EXEC measures the percentage of commands that are executable in the Virtual Home environment. This metric measures how accurately the planner can generate commands. The higher the EXEC is, the higher portion of commands generated by the LLM are usable in the environment. 

\subsubsection{Baselines}
Zero-Shot Planner \cite{huang2022language}, Prog-Prompt \cite{singh2022progprompt}, and Tree-Planner \cite{hu2023treeplanner} are three prevalent OpenAI-API-based task-planning methods with strong performance. For experiments without error correction, we compare our method to these three methods. For experiments with error correction, we compare our method with the Prog-Prompt \cite{singh2022progprompt} and Tree-Planner \cite{hu2023treeplanner} method tested in the work of Tree-Planner with Global Replan methods. We consider these aforementioned methods as our baseline. The results of Zero-Shot Planner, Tree-Planner, and Iterative-Planner are directly quoted from Tree-Planner \cite{hu2023treeplanner} since the code is inaccessible online. 

\subsubsection{Implementation Details}
Our experiments use the OpenAI GPT-3.5 \cite{brown2020language} model API as the LLM backbone across the evaluated methods. In the experimentation of our method, we used the backbone of Prog-Prompt and GPT-3.5 to generate our plans, during which we selected 4 representative plans from the dataset as in-context learning examples: ``put the wine glass in the kitchen cabinet'', ``wash mug'', ``wash clothes'', and ``put apple in fridge''. To generate diverse plans, we adopted the temperature to be 0.1, with the number of generated plans set to 30 during the plan generation phase. During the plan reordering phase, we adopted the temperature to be 0.65, with the number of plans set to 20 to generate diverse plans. 


\subsection{Experimental Results}

\textbf{Main results.} In Table \ref{tab:without_correction_table}, we presented the result of experiments without correction. These experiments are run ten times on the unseen test set in the simulator, then calculate each evaluation metric's mean and standard deviation. Noticeably, our method outperforms the previous state-of-the-art by around 10\% in terms of SR. Our method also achieves higher GCR and comparable executability, which means our method is capable of completing more sub-goals and generating more executable plans. Moreover, after adding the error correction module as shown in Table \ref{tab:with_correction_table}, our performance steadily improves overall metrics and surpasses other baseline methods in terms of SR and GCR. For executability, our method slightly underperforms Prog-Prompt. We believe this might be due to the length of the plans generated by the two methods. Prog-Prompt generates longer and more redundant plans compared to our method, as discussed in detail in \ref{analysis}. These redundant commands are mostly executable, which could account for the slight increase in executability for Prog-Prompt.

\textbf{Ablation of different node selection methods.} During the execution process, the agent can either randomly select a child node from the tree or choose the child node with the maximum votes as proposed in Vote-Tree-Planner. To demonstrate the effectiveness of our voting design, we compared the performance between these selection methods in Table \ref{tab:abalation_table}. The voting mechanism guides the planner toward correct executions and significantly enhances the overall performance.

\subsection{Qualitative Analysis}
\label{analysis}
In this section, we discuss scenarios, where our approach demonstrated better handling of cases and reduced the length of the plans compared to those generated by Prog-Prompt \cite{singh2022progprompt}.

\subsubsection{Failure cases in Prog-Prompt}
For example, in the case ``Microwave Salmon'', the plans provided by Prog-Prompt and our method are shown on the left in Figure \ref{fig:comparison}. Both of the methods contain some necessary commands in green and our method generates commands that are correct and more concise. For example, the plan from Prog-Prompt tries to find the microwave twice, while our method only tries to do it once. Also, the plan from Prog-Prompt tries to switch on the microwave when the microwave is still open, while our method handles this correctly. 

\subsubsection{Redundancy cases in Prog-Prompt}
For example, in the case ``Put Salmon In The Fridge'', the plans provided by Prog-Prompt and our method are shown on the right in Figure \ref{fig:comparison}. Both of the methods generate the necessary commands for getting the salmon and finding the fridge. However, the plan generated by Prog-Prompt has many more condition checks and even unnecessary commands. For example, the last three commands from Prog-Prompt intend to close the fridge, but instead of doing \textit{assert(`fridge' is `closed’), else: close(`fridge’)}, the plan does \textit{assert(`fridge' is `opened’), else: open(`fridge’), close(`fridge’)}, hence the plan may result in redundant actions, i.e. opening the fridge then close it even if the fridge is originally closed. On the other hand, our method generates a more concise plan without redundant commands.

From these examples, we can see that Prog-Prompt generates plans with more redundancy and has to query the LLM during each assertion phase, leading to significantly higher token consumption. In contrast, our method avoids querying the LLM and instead makes corrections based on the planning tree.

\section{Conclusion} 
In this paper, we introduced Vote-Tree-Planner, a novel strategy for task planning that leverages the capability of Large Language Models (LLMs). Vote-Tree-Planner effectively addresses the instability inherent in repetitive planning commands and reduces unnecessary assertion queries during plan execution. Our experiments conducted in the Virtual Home simulation environment indicated that our approach outperforms baseline methods, achieving new state-of-the-art performance with higher success rates, improved goal condition recall, and comparable executability. Furthermore, the plans generated by our strategy are notably shorter and exhibit less command repetition. We contend that Vote-Tree-Planner establishes a new benchmark in LLM-based task planning by improving both query efficiency and performance. We anticipate that our contributions will inspire continued research and development within this field.

\textbf{Limitations.} Despite achieving state-of-the-art benchmarks, our method exhibits limitations. The efficacy of our pipeline remains heavily contingent upon the capabilities of Large Language Models (LLMs), and it is susceptible to variability due to the inherent randomness in LLM outputs, albeit with enhanced stability.

\textbf{Future Work.} We aim to refine the integration of LLMs within the planning correction process to ensure that plans remain relevant and adaptive to changes in the environment. Additionally, improving token efficiency remains a critical objective, which could lead to more streamlined interactions and reduced computational demands. These advancements will not only enhance the robustness of our method but also extend its applicability and effectiveness in dynamic settings. Last but not least, different LLM models possess different levels of capabilities. Comparing planners with different LLM backbones (i.e. Llama \cite{touvron2023llama}, GPT-4 \cite{achiam2023gpt}, Gemini \cite{team2023gemini}) is also one of our future work directions.


\bibliographystyle{plainnat}
\bibliography{references}

\end{document}